\documentclass{article}
\usepackage{graphicx} 
\usepackage{parskip}
\usepackage{natbib}
\usepackage{float}
\usepackage{url}
\usepackage{iclr2024_conference,times}
\usepackage{listings}
\usepackage{xcolor} 
\usepackage{hyperref}

\definecolor{codegray}{gray}{0.95}  

\iclrfinalcopy
\title{Leafy Spurge Dataset: \\ Real-world weed classification \\ within aerial drone imagery}

\author{%
Kyle Doherty \\
MPG Ranch \\
\texttt{kdoherty@mpgranch.com}
\And
Max Gurinas \\
University of Chicago Laboratory Schools \\
\texttt{maxgurinas@gmail.com}
\And
Erik Samsoe \\
MPG Ranch \\
\texttt{esamsoe@mpgranch.com}
\And
Charles Casper \\
MPG Ranch \\
\texttt{ccasper@mpgranch.com}
\And
Beau Larkin \\
MPG Ranch \\
\texttt{blarkin@mpgranch.com}
\And
Philip Ramsey \\
MPG Ranch \\
\texttt{pramsey@mpgranch.com}
\And
Brandon Trabucco \\
Carnegie Mellon University \\
\texttt{brandon@btrabucco.com}
\And
Ruslan Salakhutdinov \\
Carnegie Mellon University \\
\texttt{rsalakhu@cs.cmu.edu}
}

\begin{document}

\maketitle
\newpage

\begin{abstract}
Invasive plant species are detrimental to the ecology of both agricultural and wildland areas. \textit{Euphorbia esula}, or leafy spurge, is one such plant that has spread through much of North America from Eastern Europe. When paired with contemporary computer vision systems, unmanned aerial vehicles, or drones, offer the means to track expansion of problem plants, such as leafy spurge, and improve chances of controlling these weeds. We gathered a dataset of leafy spurge presence and absence in grasslands of western Montana, USA, then surveyed these areas with a commercial drone. We trained image classifiers on these data, and our best performing model, a pre-trained DINOv2 vision transformer, identified leafy spurge with 0.84 accuracy (test set). This result indicates that classification of leafy spurge is tractable, but not solved. We release this unique dataset of labelled and unlabelled, aerial drone imagery for the machine learning community to explore. Improving classification performance of leafy spurge would benefit the fields of ecology, conservation, and remote sensing alike. Code and data are available at: { \href{https://leafy-spurge-dataset.github.io}{leafy-spurge-dataset.github.io}}.
\end{abstract}

\section{Introduction}
Gathering ecological data requires expertise in species identification, spatial planning, and haste to capture ephemeral patterns. Consequently, scaling ecological monitoring efforts across larger areas is difficult without supporting technologies such as satellite or drone-based imaging \citep{turner2005remote}. Agriculture has pioneered these remote sensing approaches to detect and map changes to crop health \citep{mulla2013twenty} and weed plant occurrence \citep{lamb2001weed}. Increasingly, remote sensing tools are applied in wildland settings, to monitor change in plant communities after ecological disturbance and subsequent management responses \citep{pettorelli2014satellite}. While the challenges posed by large-scale ecological monitoring are formidable, this effort is critical to tracking change in natural ecosystems.

Many parallels exist between precision agriculture and ecological applications of remote sensing, but there are notable differences. The diversity of species present in wildland systems is often far greater than those in agricultural contexts \citep{phalan2011minimising}. Thus, for tasks, such as image classification, discerning target plants from background can be more difficult as the background domain is more varied. Similarly, the terrain of wildlands is more complex than in agricultural sites, which are situated in flat areas to facilitate mechanized tillage, seeding, and pest management \citep{ozkan2020sitesuitability}.  Complex terrain generates varied lighting conditions \citep{corripio2003vectorial}, which may pose challenges for classifiers \citep{rodriguezgaliano2012landcover}. Furthermore, diagnostic features of target plants, such as flower and leaf morphology, may be impossible to resolve in coarse-grained images, necessitating the use of drone-based platforms that can fly lower and gather fine-grained information \citep{amputu2023uas}. 

\begin{figure}[t]
  \centering
  \includegraphics[width=0.8\linewidth]{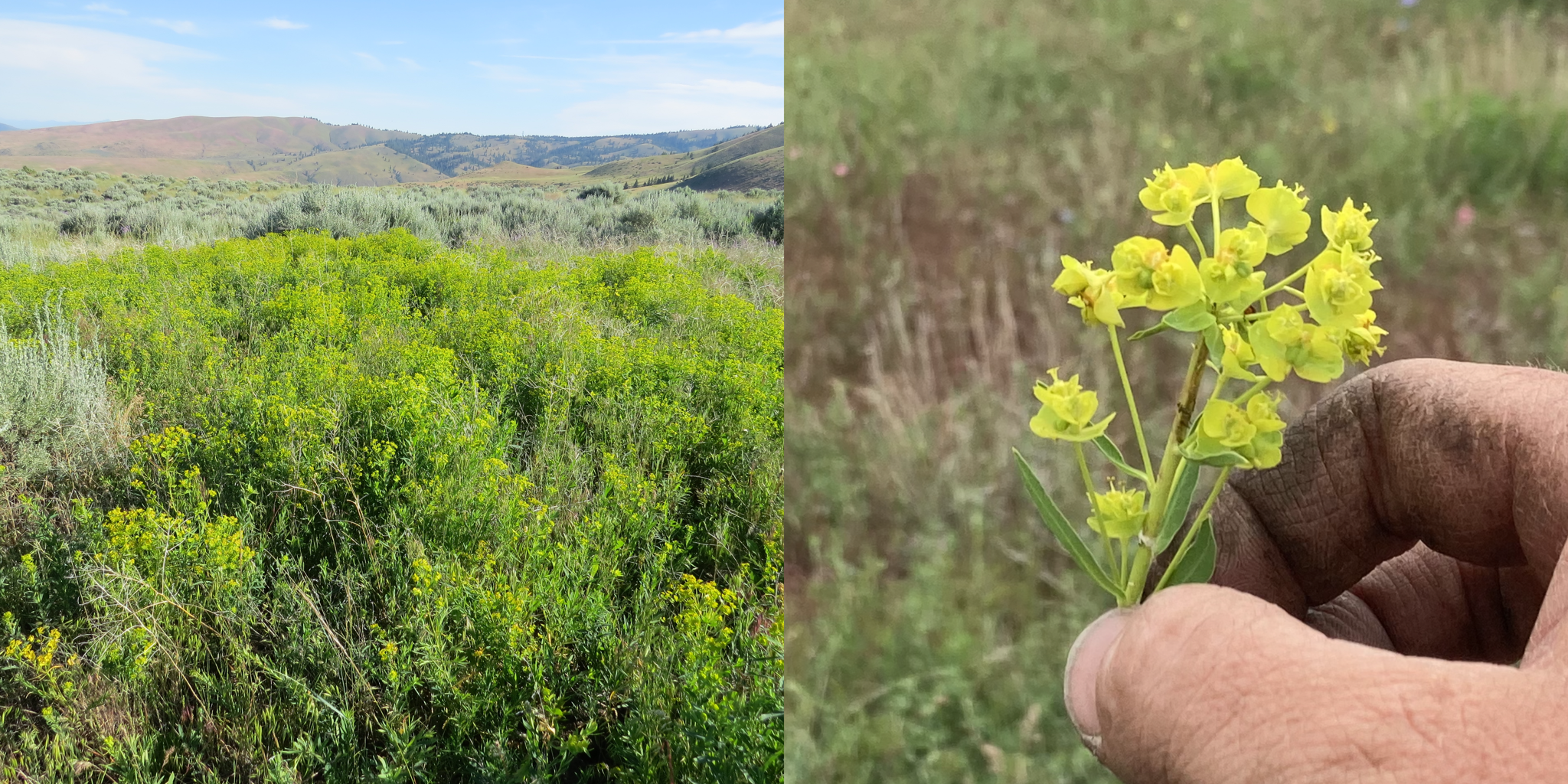}
  \caption{A sagebrush community invaded by leafy spurge (left) and a closeup of a leafy spurge inflorescence (right).}
  \label{fig:spurge_photo}
\end{figure}

One important use-case of remote sensing is that of weed plant detection. When an invasive plant invades a natural area, it can cause harm by a variety of mechanisms, including competition with native plants for space and resources \citep{maron2008field}, disruption of pollinator services \citep{pearson2012darwin}, catalyzing catastrophic wildfire regimes \citep{balch2013annualgrass,bradley2018cheatgrass}, and others. Once established, invasive plants are difficult to remove, requiring expensive monitoring and treatment. Thus, early warning systems capable of detecting invasion by weed plants are of great value to preserving ecological integrity of natural areas naïve to invasion and those treated for past invasions \citep{lass2005review}. When paired with computer vision systems, drones promise an agile means to detect weed plants and rapidly respond to incipient invasion in wildlands, but the robustness of weed classifiers in these complex areas is under-explored.

Leafy spurge (\textit{Euphorbia esula}; \textbf{Fig.~\ref{fig:spurge_photo}}) is a problem weed introduced to North America in the late 19th century, likely in agricultural equipment or crop seeds imported from Eastern Europe \citep{dunn1985origins}. Leafy spurge expanded in agricultural areas then escaped into wildlands. The plant is noxious to cattle and wild grazing mammals, which avoid infested areas, leading to economic losses \citep{bangsund1993economic}. Chemical, biological, and cultural control methods were implemented extensively to manage and limit the spread of leafy spurge, to preserve the integrity of wildlands \citep{gaskin2021managing}. Yet, for most spurge-invaded regions, little is known about the rate of spread of this plant or the efficacy of management actions aimed at controlling it.

We gathered high resolution drone imagery of grasslands undergoing ecological restoration in western Montana, USA where leafy spurge has established and is targeted for removal. In parallel we collected ground truth of spurge presence and absence with precision GPS systems throughout the study region. We then evaluated performance of classifiers tasked with identifying leafy spurge in our aerial imagery, and report the first benchmarks on this objective below. We release these data here for the purpose of advancing spurge detection and management as well as furnishing the machine learning research community with a unique, real-world dataset suitable for testing state of the art computer vision systems. We discuss the novel aspects of our data below and suggest research directions, such as study of few-shot learning, for which it is ideally suited.

\section{Image acquisition and post-processing}

\begin{figure}[h]
  \centering
  \includegraphics[width=0.8\linewidth]{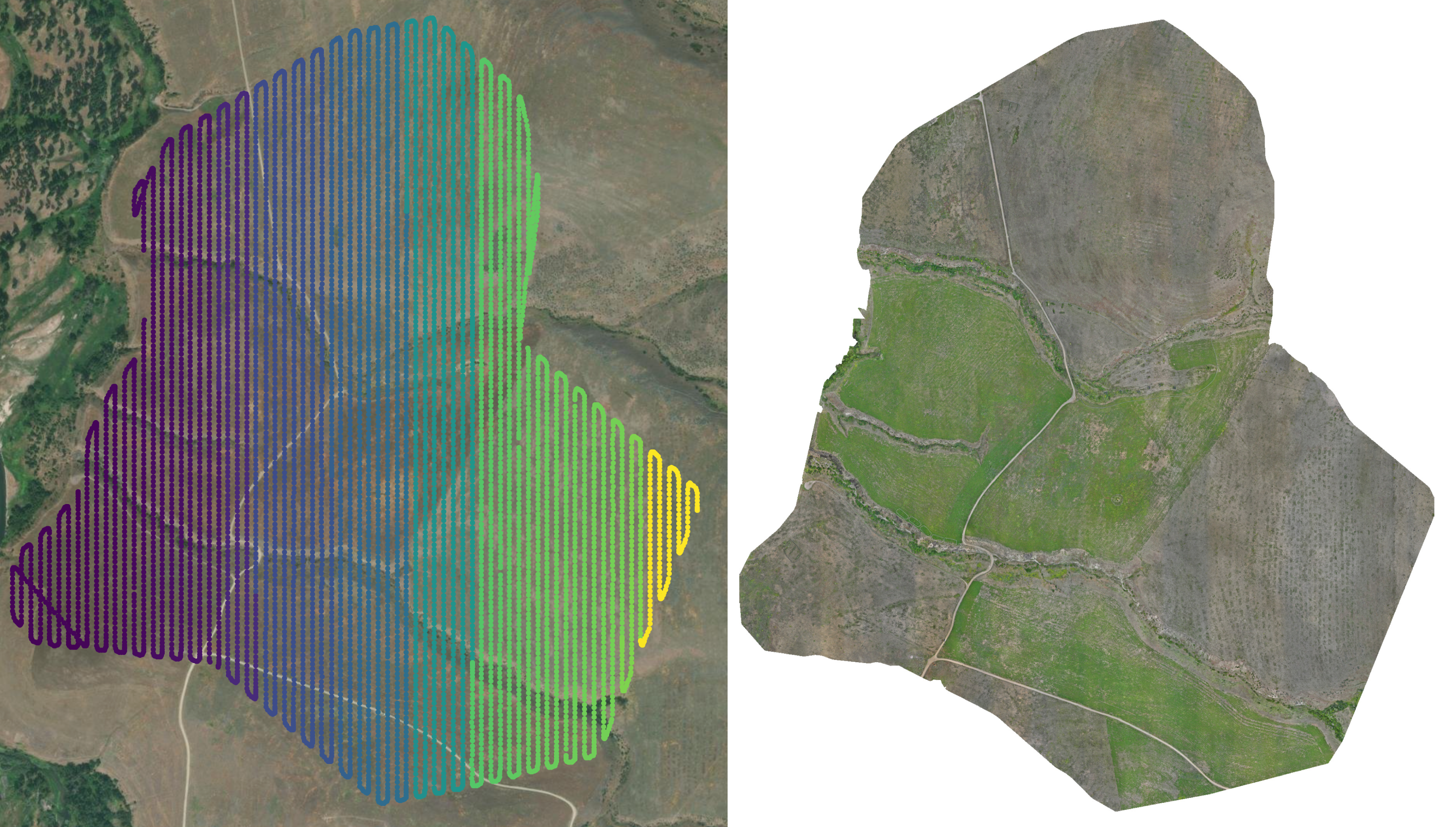}
  \caption{\small Left: A map of the drone survey flight plan where each point represents the coordinates of a photo in the survey, colored by time of photo. Right: These overlapping images are processed into a large, georeferenced orthomosaic (right).}
  \label{fig:flight}
\end{figure}

We surveyed the study area on June 12, 2023, during a 67-minute window (from 12:32 to 13:39) with a DJI Mavic 3M drone. The drone captured 8241 images at 50 m above ground level across an area of 118 hectares (\textbf{Fig.~\ref{fig:flight}}).  During the survey there was light wind and sparse cloud cover at 3700 m. We programmed the drone flight such that images overlapped to improve performance of the feature matching algorithm during post-processing, which merges raw images into a single spatially contiguous and georeferenced image, or orthomosaic.  The side overlap ratio was 70\% and a front overlap ratio was 80\% between adjacent images. This process generated an estimated ground sampling distance (GSD) of 1.27 cm per pixel in the resultant orthomosaic. A key benefit of orthosmosaicing imagery is to georeference each pixel (as opposed to only centroids of single images), which enables predictive mapping of the target weed with downstream classifiers. 

We performed all post-processing with the DroneDeploy software suite (\url{https://www.dronedeploy.com}). This involved feature matching of overlapping images, correction of geometric distortions, and the generation of a georeferenced orthomosaic. Prior to surveying, we established 32 ground control points (GCPs; points for which positions were verified with a precision GPS) across the study area. These GCPs were used during post-processing to further minimize georeferencing error of pixels in the orthomosaic product. The mean Root Mean Squared Error (RMSE) of GCP position was 7.32 cm after post-processing. 
 
\section{Ground truth acquisition}

\begin{figure}[t]
  \centering
  \includegraphics[width=0.5\linewidth, height=159pt]{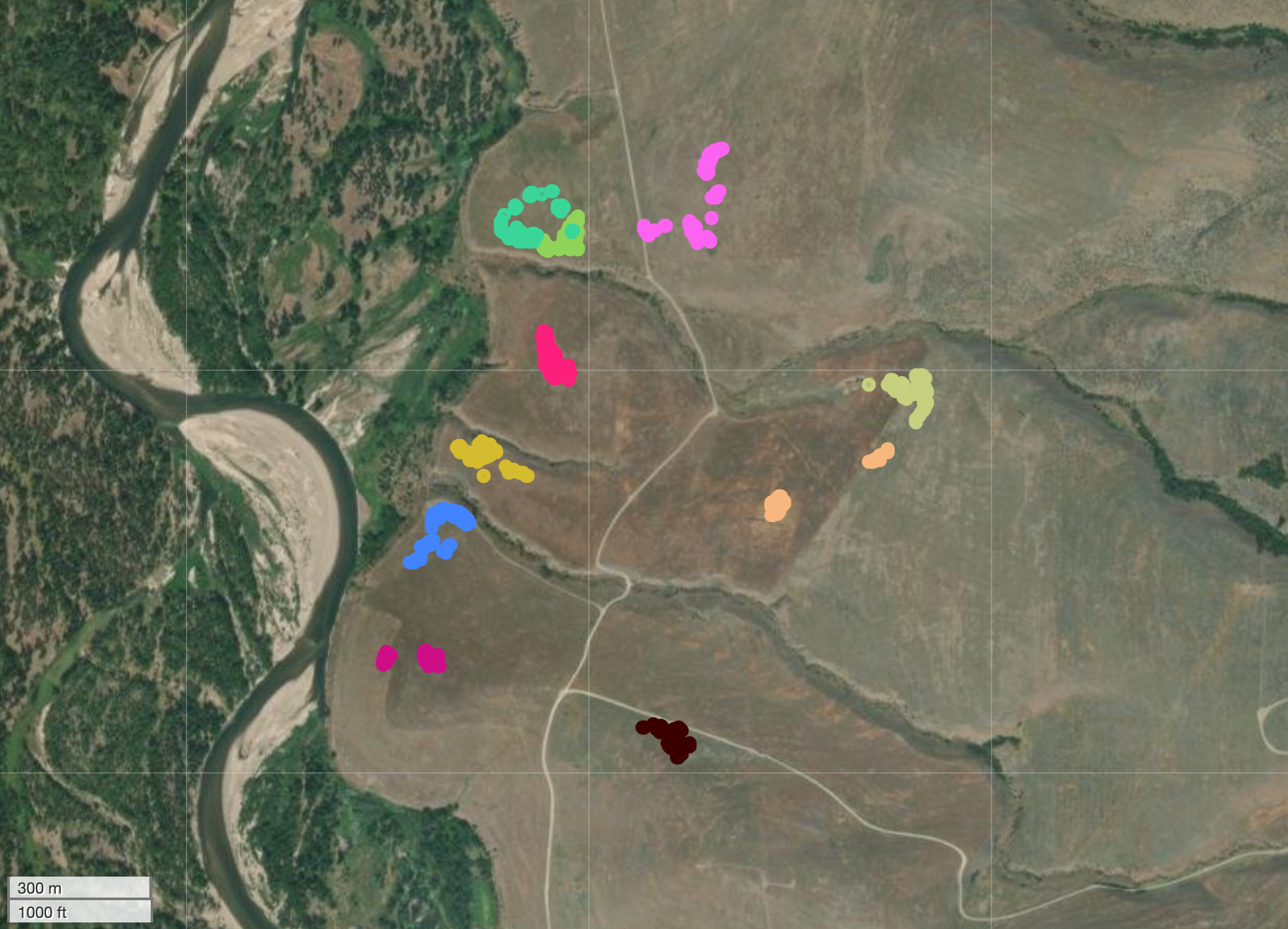}
  \hspace{0.1in}
   \includegraphics[width=0.4\linewidth]{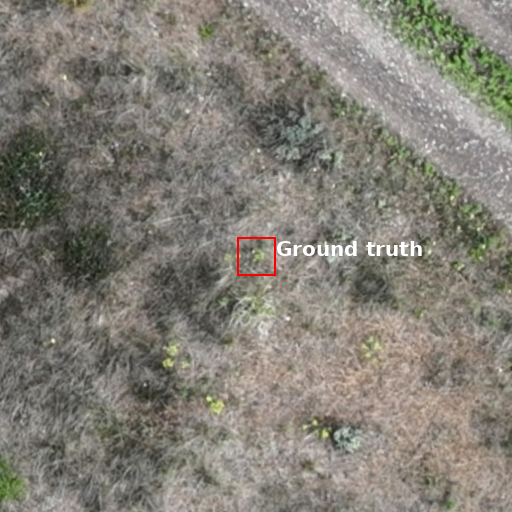}
  \caption{\small Left: Locations of leafy spurge ground truth samples. Point colors indicate sampling site identity.
  Right: The red box indicates the dimensions of a leafy spurge ground truth sample (0.5 x 0.5m).}
  \label{fig:sampling_locations}
\end{figure}

After surveying our study area botanists visited sites within to gather ground truth of spurge presence and absence. Upon visiting a site, botanists conducted random walks to gather coordinates of spurge presence and absence using an Emlid RS2 GPS, capable of reporting positions with sub-centimeter accuracy. When a technician encountered a target plant, they would record its position. Technicians also gathered coordinates for spurge absences in this manner. For spurge absence cases, ground truth indicates that no spurge plants were detected in a 0.5 x 0.5m box centered on the coordinates. During the walk technicians gathered data until they acquired 50 presences and 50 absences at each site. We visited a total of 10 sites, sampling in this manner, and accumulated 500 examples per presence/absence class (\textbf{Fig.~\ref{fig:sampling_locations}}; left panel). While the majority of sites were geographically separated, two sites overlapped due to weather-related time constraints. 


We extracted training images from crops of our orthoimage corresponding to ground truth bounding boxes as well as a larger size (described below) to use in downstream classification. Each instance of ground truth corresponds to 0.5 x 0.5m, or 39 x 39 pixels (\textbf{Fig.~\ref{fig:sampling_locations}}; right panel).

\section{Benchmarks of Leafy Spurge Classifiers}

We evaluated two vision backbone architectures for the image classification task: ResNet50 \citep{he2016deep} and DINOv2 \citep{oquab2023DINOv2}. ResNet50 is a widely adopted convolutional neural network, while DINOv2 is a more recent vision transformer-based model. We used pre-trained weights from their respective checkpoints (\verb|facebook/DINOv2-base| for DINOv2) for initialization. We preprocessed the images by applying the standard ImageNet normalization and resizing to 224x224 pixels. We conducted experiments with two image sizes from the dataset: 39x39 pixels ('crop' revision) and 1024x1024 pixels ('context' revision). The intent of training with larger, 1024 pixel images was to explore if broader context around the ground truth could aid classifier performance, as is reported for larger vision transformer models \citep{mm1-methods-analysis-insights}. To enhance model generalization and mitigate overfitting, we applied the following data augmentation techniques during training with a probability of 0.5: ColorJitter (brightness=0.8, contrast=0.7, saturation=0, hue=0), RandomHorizontalFlip, RandomVerticalFlip, and RandomRotation (degrees=90).

We trained the models for 1000 epochs using the AdamW optimizer \citep{loshchilov2019decoupled} with a learning rate of 1e-5 for ResNet50 and 2.5e-6 for DINOv2. We employed a batch size of 24  with appropriate random seeding for reproducibility. For each experiment (combination of model and image size, and dataset revision), we used 5 unique seeds to account for variability. For each seed, we split the training set into 80\% for training and 20\% for validation. We evaluated model performance on both the validation and test sets during training. For our performance metric we calculated the 95\% confidence interval of the proportion of correctly classified samples.

\section{Results}

\begin{figure}[t]
  \centering
  \includegraphics[width=\linewidth]{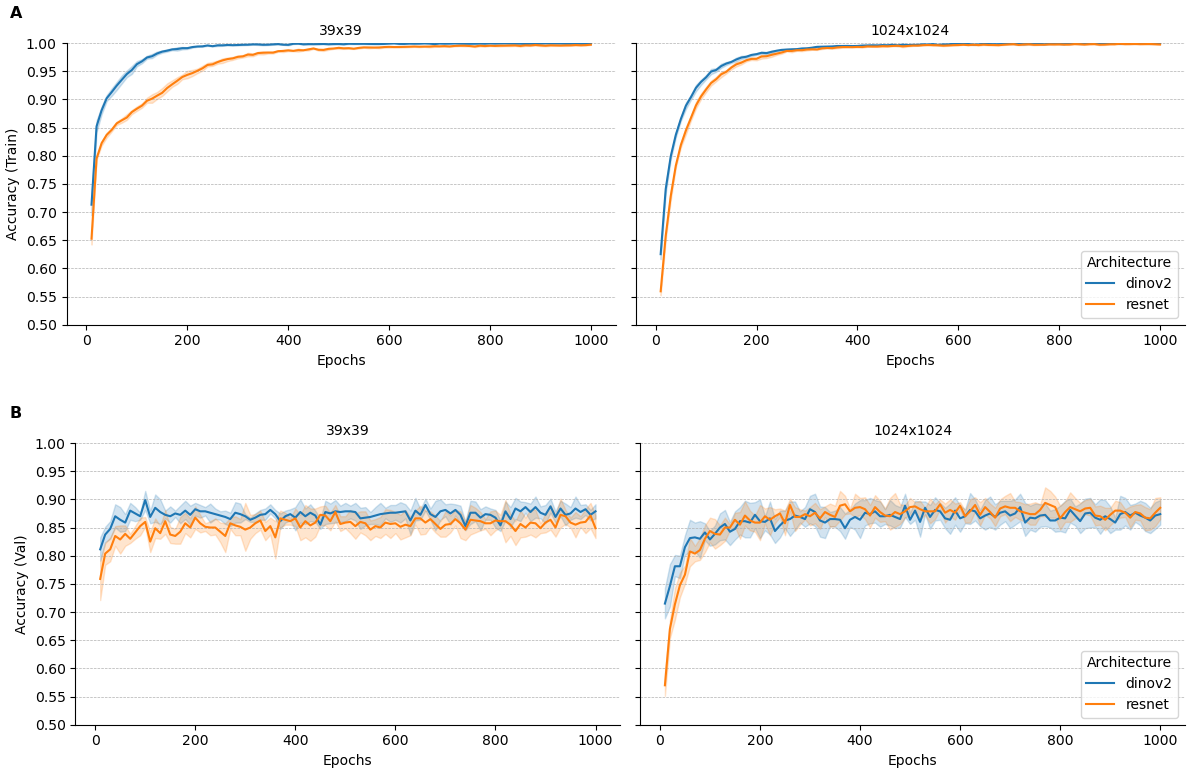}  
 \vspace{-0.2in}
  \caption{\small We present classification results for the Leafy Spurge dataset when training on 39x39 pixel (left panels) and 1024x1024 pixel (right panels) images. We tested two architectures: ResNet50 (orange) and DINOv2 (blue). We report training (A) and validation (B) set accuracy. We represent cross-validated means as lines and 95\% confidence intervals as bands.}
  \label{fig:accuracy}
\end{figure}

\begin{table}[t]
    \centering
    \begin{tabular}{|l|l|l|}
    \hline
    \textbf{Architecture} & \textbf{Image Size} & \textbf{Test Accuracy} \\
    \hline
    DINOv2 & 39x39 & 0.84 (0.82, 0.86) \\
    ResNet50 & 39x39 & 0.83 (0.80, 0.86) \\
    DINOv2 & 1024x1024 & 0.74 (0.72, 0.75) \\
    ResNet50 & 1024x1024 & 0.66 (0.64, 0.67) \\
    \hline
    \end{tabular}
    \caption{\small Test accuracy of ResNet50 and DINOv2 architectures when trained on 39x39 or 1024x1024 pixel images. We report 95\% confidence intervals of accuracy metrics in parentheses.}
    \label{table:test_accuracy}
\end{table}

We found both DINOv2 and ResNet50 architectures suitable for detection of the target plant, leafy spurge (\textbf{Table~\ref{table:test_accuracy}}). Notably, models trained on smaller images (39x39) whose dimensions correspond directly to the 0.5x0.5m bounds of ground truth performed better than those trained on larger images (1024x1024). This diverges from our expectation that the vision transformer model, DINOv2, would benefit from higher resolution images. It is not clear if the drop in performance relates to difference in image size directly, because it could be the case that expanding the bounds of the image beyond that of the ground truth could lead to accidental inclusion of the target plant in negative examples. While larger images performed worse than smaller ones, it was the case that DINOv2 outperformed the ResNet50 for larger images by a wide margin (0.72 DINOv2 versus 0.64 ResNet50). Furthermore, at both images sizes, DINOv2 appeared to converge more quickly during training (\textbf{Fig.~\ref{fig:accuracy}}).  

\section{Applied considerations for future study}

We hope that those exploring our data will tailor their work to benefit the land management community whose objective is to contain and remove leafy spurge. One primary consideration is that of spatial scale of spurge predictions, where managers would benefit from more granular predictions. As mentioned earlier, our ground truth data and the size of the smaller images for classifier training were 0.5 x 0.5m, which corresponds to roughly the size of a large leafy spurge plant. Mapping spurge extents by tiling out the orthomosaic into thumbnails of this size, conducting inference on tiles, and reconstituting the products into a mosaic would indeed be useful to land managers. However, pixel-level inference could also prove useful for automated drone spray systems that are currently under development. Pixel-level products could potentially be used to treat leafy spurge selectively, minimizing harm to other plants. While gathering pixel-level ground truth is not feasible presently, we might supply larger scenes from an orthomosaic in conjunction with ground truth bounding boxes as prompts to Segment Anything \citep{kirillov2023segment}, or similar foundation models capable of image segmentation from less granular information. 

In addition to labelled images, we also serve the full orthomosaic, excluding test regions. The vast majority of surveyed areas were not visited by ground truth teams. These unlabelled data could be used in an unsupervised pipeline to further enhance downstream presence/absence classifier performance. Successful demonstrations of unsupervised learning applications for aerial imagery would greatly benefit spurge control efforts as well as the field of remote sensing generally, where vast amounts of unlabelled aerial images are a common condition.

\section{Relevance to study of foundation models}

Our data are also well-suited for the study of generative models, specifically, for exploration of large models in zero and few-shot contexts. A key challenge in evaluating these so-called foundation models is that they are trained on large bodies of data which are often public-facing. Thus, zero and few-shot researchers may struggle to source data that are truly outside the domain of the training set. For example, common semantic concepts, such as mammal species, are well represented in the corpus of the text-to-image diffusion model, Stable Diffusion \citep{rombach2022highresolution}, and naive evaluations of zero and few-shot generations of these concepts may not be fair due to data leakage \citep{trabucco2023effective}. 

In contrast, our data release represent the first instance of top-down images of leafy spurge imaged from 50m above ground level \citep{trabucco2023effective}. Researchers can study our data with confidence that it lies outside the training domains of large generative models released prior to this publication.

\section{Conclusions}

The Leafy Spurge Dataset offers researchers a unique opportunity to advance the fields of ecology, remote sensing, and machine learning in parallel. In our initial analyses, we found that the task of leafy spurge presence/absence classification is tractable, but not solved. In future work we hope to explore performance benefits when incorporating diverse modalities of data offered by drone surveys, including structure-from-motion elevation products and multi-spectral images, which could be paired with standard three-channel data. 

\section{Data Availability}
Our data are hosted publicly as a Hugging Face Dataset (\url{https://www.huggingface.com}). We serve two image sizes corresponding to the 39x39 and 1024x1024 pixel images as configs "crop" and "context", respectively. Additionally, we serve the full unlabelled orthomosaic as config "unlabelled."

\begin{lstlisting}[language=Python, caption=Python code for loading our dataset from Hugging Face]
from datasets import load_dataset

#39x39 pixel train and test data
crop_train = load_dataset('mpg-ranch/leafy_spurge', 
    'crop', split='train') 
crop_test = load_dataset('mpg-ranch/leafy_spurge',
    'crop', split='test') 

#1024x1024 pixel train test data
context_train = load_dataset('mpg-ranch/leafy_spurge',  
    'context', split='train') 
context_test = load_dataset('mpg-ranch/leafy_spurge', 
    'context', split='test') 

#unlabelled orthomosaic
orthomosaic = load_dataset('mpg-ranch/leafy_spurge',
    'unlabelled', split='train') 
\end{lstlisting}

\section{Acknowledgements}
We would like to thank the staff at MPG Ranch for assistance in gathering ground truth and aerial imagery, as well as the owners of MPG Ranch for their ongoing support of applied ecological, conservation, and machine learning research. 

\bibliographystyle{apalike}
\bibliography{references}

\section{Appendix A. Drone and Sensor Specifications}

The Mavic 3M is equipped with Real-Time Kinematic (RTK) positioning to enhance the GPS accuracy and ensure centimeter-level precision in the camera position. The drone features a 4/3 CMOS image sensor, with a resolution of 20 Megapixels (MP) and operated within an RGB color space. The lens provided a field of view (FOV) of 84° and an equivalent focal length of 24 mm. The ISO range was set between 100 and 6400, with a median shutter speed of 1/640 s. Additional specifications include an actual focal length of 12.29 mm, an aperture of f/2.8, and a minimum exposure time of 1/2,000 s. Images, with dimensions of 5280 X 3956 pixels, varied in size from 9.4 MB to 12.6 MB.

\end{document}